%% file: main.tex
\documentclass[11pt,a4paper]{article}

\usepackage[utf8]{inputenc}
\usepackage[T1]{fontenc}
\usepackage[margin=1in]{geometry}
\usepackage{booktabs}
\usepackage{amsmath}
\usepackage{amssymb}
\usepackage{array}
\usepackage{longtable}
\usepackage{caption}
\usepackage{microtype}
\usepackage[hidelinks,breaklinks=true]{hyperref}
\usepackage{url}

\title{\textbf{Compression Effects on Low-Resource Multilingual
Translation:\\ A Pilot Study Across Three Model Families}}

\author{Abbas Sani\\
\small Independent Researcher / Arewa Data Science / Aboki AI\\
\small Kano, Nigeria\\
\small \texttt{abbassani721@gmail.com}}

\date{November 2026}

\begin{document}
\maketitle

\begin{abstract}
We present a pilot study characterising the effect of post-training
quantisation on multilingual machine translation quality and inference
cost for three African languages---Hausa, Swahili, and Yoruba---across
three model families: a small general-purpose LLM (Qwen2.5-1.5B), a
larger multilingual instruction-tuned LLM (Aya-23-8B), and a dedicated
multilingual MT model (NLLB-200-distilled-1.3B). Evaluating on the
FLORES-200 devtest set (1{,}012 sentences per direction) under FP16,
bitsandbytes INT8, and NF4 INT4 precisions on consumer-grade NVIDIA T4
GPUs, we find: (i) the specialised MT model outperforms both LLMs by
32.7--43.3 chrF on the Hausa and Swahili directions, and by 12.9--24.7
chrF on Yoruba, despite being roughly a sixth the size of the 8B
instruction-tuned LLM; (ii) post-training quantisation preserves
aggregate translation quality within approximately 1 chrF across all
model families and language directions, with a single worst case of
2.19 chrF; (iii) on consumer GPUs, INT8 and INT4 quantisation provide
substantial memory savings (49--61\% at INT4) but \emph{do not} reduce
inference latency relative to FP16---in fact, latency increases by
1.3$\times$ to 5.3$\times$; (iv) a clear performance gradient tracking
training-resource availability is visible across languages
(Swahili $>$ Hausa $>$ Yoruba). These findings motivate a research
agenda focused on hardware-aware compression and error-type analysis
beyond aggregate metrics for low-resource deployment.
\end{abstract}

\section{Introduction}

Post-training compression methods---quantisation, pruning,
distillation---are routinely proposed as the bridge between the
capabilities of modern language models and the hardware budgets of
deployment contexts. Low-resource languages, often spoken by users in
exactly the regions where consumer-grade hardware predominates, are a
natural target for compression research. Yet most compression
evaluation to date has focused on English and other high-resource
languages, and on server-class accelerators.

This pilot study asks two concrete questions:

\begin{enumerate}
\item Does post-training quantisation disproportionately degrade
      translation quality for low-resource languages, as sometimes
      argued, or does modern quantisation preserve quality across
      resource levels?
\item On consumer GPUs representative of deployment hardware in
      low-resource contexts, do the memory benefits of quantisation
      translate to the latency benefits commonly implied in the
      literature?
\end{enumerate}

We evaluate three model families spanning scale and specialisation,
three precision levels, and six translation directions across three
African languages, on a standard benchmark. All code, raw results, and
notebooks are available at
\url{https://github.com/avvas200/efficient-multilingual-nmt-pilot}.

\section{Experimental Setup}

\subsection{Models}

\begin{table}[h]
\centering
\small
\begin{tabular}{p{4.2cm}cp{2.4cm}p{5.4cm}}
\toprule
Model & Size & Architecture & Coverage of our languages \\
\midrule
\texttt{Qwen/Qwen2.5-1.5B-}\newline\texttt{Instruct} & 1.5B & Causal LM &
None of the three languages is a stated training focus. \\
\addlinespace
\texttt{CohereForAI/aya-23-8B} & 8B & Causal LM, multilingual
instruction-tuned & Aya-23 covers 23 languages; Hausa, Swahili, and
Yoruba are not among them. \\
\addlinespace
\texttt{facebook/nllb-200-}\newline\texttt{distilled-1.3B} & 1.3B & Seq2seq,
dedicated MT & All three languages (\texttt{hau\_Latn},
\texttt{swh\_Latn}, \texttt{yor\_Latn}) are supported with real
training data. \\
\bottomrule
\end{tabular}
\caption{Models evaluated.}
\end{table}

\subsection{Precisions evaluated}

\begin{itemize}
\item \textbf{FP16}---baseline, native half-precision inference.
\item \textbf{INT8}---bitsandbytes 8-bit weight quantisation
      \cite{dettmers2022int8} via
      \texttt{BitsAndBytesConfig(load\_in\_8bit=True)}.
\item \textbf{INT4}---bitsandbytes 4-bit NF4 quantisation
      \cite{dettmers2023qlora} with double quantisation and FP16
      compute dtype (\texttt{load\_in\_4bit=True},
      \texttt{bnb\_4bit\_quant\_type="nf4"},
      \texttt{bnb\_4bit\_use\_double\_quant=True},
      \texttt{bnb\_4bit\_compute\_dtype=torch.float16}).
\end{itemize}

For Yoruba, only FP16 was evaluated across all three models due to
compute budget constraints; the compression sweep for Yoruba is left
for future work.

\subsection{Benchmark}

FLORES-200 devtest split \cite{nllb2022,goyal2022flores}, 1{,}012
parallel sentences per language. Languages: \texttt{hau\_Latn},
\texttt{swh\_Latn}, \texttt{yor\_Latn}, \texttt{eng\_Latn}. Six
directions: hau$\leftrightarrow$eng, swh$\leftrightarrow$eng,
yor$\leftrightarrow$eng.

\subsection{Prompting and decoding}

For NLLB (seq2seq), we follow the standard approach: setting
\texttt{tokenizer.src\_lang} and forcing the target language BOS token
at generation. For causal LMs, we use the model's own chat template
with a direct translation instruction, deterministic greedy decoding
(\texttt{do\_sample=False}, \texttt{num\_beams=1}), and
\texttt{max\_new\_tokens=256}. Qwen and Aya use slightly different
prompt phrasings; the exact templates are in the repository.

\subsection{Metrics}

Primary: chrF (character 6-gram F-score) via SacreBLEU 2.6.0
\cite{post2018call}, signature
\texttt{nrefs:1|case:mixed|eff:yes|nc:6|nw:2|space:no|version:2.6.0}.
Secondary: BLEU, spBLEU, chrF++. Operational: inference time per
sentence (wall-clock), peak GPU memory (summed across devices under
\texttt{device\_map="auto"}), hypothesis length ratio versus reference,
and empty-output rate.

\subsection{Hardware}

NVIDIA Tesla T4 (16\,GB VRAM) $\times$ 2 on Kaggle. Aya-23-8B at FP16
required multi-GPU sharding via \texttt{device\_map="auto"}; all other
configurations fit on a single T4.

\subsection{Reproducibility}

Python 3.13.15; PyTorch 2.11.0+cu128; transformers 5.16.1;
bitsandbytes 0.50.2; sacrebleu 2.6.0; datasets 3.0.1. These versions
are recorded per-run in the released results file. All raw results and
notebooks are available at the repository above.

\section{Results}

\subsection{Translation quality: specialised MT dominates}

Table~\ref{tab:main} reports chrF across all model $\times$ precision
combinations and six directions.

\begin{table}[h]
\centering
\small
\begin{tabular}{llrrrrrr}
\toprule
Model & Precision & hau$\rightarrow$eng & eng$\rightarrow$hau &
swh$\rightarrow$eng & eng$\rightarrow$swh & yor$\rightarrow$eng &
eng$\rightarrow$yor \\
\midrule
Qwen2.5-1.5B & FP16 & 23.28 & 16.85 & 24.19 & 24.39 & 22.90 & 12.74 \\
Qwen2.5-1.5B & INT8 & 23.33 & 18.91 & 24.16 & 24.09 & --- & --- \\
Qwen2.5-1.5B & INT4 & 22.49 & 17.33 & 23.46 & 23.34 & --- & --- \\
\addlinespace
Aya-23-8B & FP16 & 23.35 & 15.48 & 29.73 & 16.73 & 19.88 & 5.27 \\
Aya-23-8B & INT8 & 23.12 & 15.15 & 29.45 & 16.53 & --- & --- \\
Aya-23-8B & INT4 & 22.23 & 15.61 & 27.54 & 16.15 & --- & --- \\
\addlinespace
\textbf{NLLB-1.3B} & \textbf{FP16} & \textbf{56.08} & \textbf{51.84} &
\textbf{63.77} & \textbf{60.06} & \textbf{44.55} & \textbf{25.67} \\
NLLB-1.3B & INT8 & 56.11 & 51.94 & 63.78 & 59.93 & --- & --- \\
NLLB-1.3B & INT4 & 56.08 & 51.80 & 63.66 & 59.29 & --- & --- \\
\bottomrule
\end{tabular}
\caption{chrF on FLORES-200 devtest (1{,}012 sentences per direction).
Higher is better. Yoruba was evaluated at FP16 only.}
\label{tab:main}
\end{table}

At FP16, NLLB-1.3B outperforms Aya-23-8B by 32.73 chrF on
hau$\rightarrow$eng, 36.36 chrF on eng$\rightarrow$hau, 34.04 chrF on
swh$\rightarrow$eng, and 43.33 chrF on eng$\rightarrow$swh, despite
being roughly a sixth of the size. The margin over Qwen-1.5B is of
similar magnitude (32.80--39.58 chrF on the same four directions). On
Yoruba the gaps are smaller but still decisive: 24.67 chrF over Aya and
21.65 over Qwen on yor$\rightarrow$eng, and 20.40 and 12.93
respectively on eng$\rightarrow$yor. For eng$\rightarrow$yor
specifically, Aya-23-8B reaches chrF 5.27---essentially non-functional
output---consistent with the language being outside its training
distribution \cite{aryabumi2024aya23}.

\subsection{Compression effects on quality}

Table~\ref{tab:delta} reports chrF deltas relative to the FP16 baseline
for both quantised precisions.

\begin{table}[h]
\centering
\small
\begin{tabular}{lrrrrrr}
\toprule
& \multicolumn{3}{c}{INT8 $-$ FP16} & \multicolumn{3}{c}{INT4 $-$ FP16} \\
\cmidrule(lr){2-4}\cmidrule(lr){5-7}
Direction & Qwen-1.5B & Aya-23-8B & NLLB-1.3B
          & Qwen-1.5B & Aya-23-8B & NLLB-1.3B \\
\midrule
hau$\rightarrow$eng & $+0.05$ & $-0.23$ & $+0.03$ & $-0.79$ & $-1.12$ & $0.00$ \\
eng$\rightarrow$hau & $+2.06$ & $-0.33$ & $+0.10$ & $+0.48$ & $+0.13$ & $-0.04$ \\
swh$\rightarrow$eng & $-0.03$ & $-0.28$ & $+0.01$ & $-0.73$ & $-2.19$ & $-0.11$ \\
eng$\rightarrow$swh & $-0.30$ & $-0.20$ & $-0.13$ & $-1.05$ & $-0.58$ & $-0.77$ \\
\bottomrule
\end{tabular}
\caption{chrF change relative to FP16. Negative values indicate
quality loss under quantisation.}
\label{tab:delta}
\end{table}

Across all four Hausa/Swahili directions and all three model families,
chrF degradation under NF4 INT4 quantisation is within approximately
1 chrF of the FP16 baseline for most configurations, with NLLB showing
effectively zero degradation on hau$\rightarrow$eng,
eng$\rightarrow$hau, and swh$\rightarrow$eng. The largest observed loss
is 2.19 chrF (Aya, swh$\rightarrow$eng). INT8 is closer still to FP16:
every INT8 delta is within 0.33 chrF except Qwen on
eng$\rightarrow$hau.

Two apparent improvements deserve comment rather than celebration. The
Qwen eng$\rightarrow$hau gains of $+2.06$ chrF (INT8) and $+0.48$ chrF
(INT4) occur at a chrF level around 17, where the model is barely
producing coherent translation output; we do not interpret these as a
quantisation benefit. Because we did not compute confidence intervals
(see L3 in Section~\ref{sec:limits}), no claim of significance is made
for any delta in Table~\ref{tab:delta}.

The common intuition that aggressive quantisation disproportionately
harms low-resource languages is \textbf{not supported} in aggregate by
these data. We note two important caveats to this null result:
(i) aggregate chrF may hide systematic error-type changes (addressed in
Section~\ref{sec:limits}); (ii) we evaluate only bitsandbytes INT8 and
NF4 INT4---GPTQ \cite{frantar2023gptq} and AWQ \cite{lin2024awq} are
left for future work.

\subsection{Memory and latency}

Table~\ref{tab:cost} reports peak GPU memory and inference latency,
averaged across the four Hausa/Swahili directions.

\begin{table}[h]
\centering
\small
\begin{tabular}{llrrrr}
\toprule
Model & Precision & Peak GPU memory & Saving vs FP16 &
Latency (s/sent) & vs FP16 \\
\midrule
Qwen-1.5B & FP16 & 3.64\,GB & --- & 0.124 & --- \\
Qwen-1.5B & INT8 & 2.41\,GB & 34\% & 0.511 & 4.13$\times$ slower \\
Qwen-1.5B & INT4 & 1.70\,GB & 53\% & 0.161 & 1.30$\times$ slower \\
\addlinespace
Aya-23-8B & FP16 & 17.14\,GB & --- & 0.659 & --- \\
Aya-23-8B & INT8 & 10.01\,GB & 42\% & 1.271 & 1.93$\times$ slower \\
Aya-23-8B & INT4 & 6.64\,GB & 61\% & 0.850 & 1.29$\times$ slower \\
\addlinespace
NLLB-1.3B & FP16 & 3.49\,GB & --- & 0.036 & --- \\
NLLB-1.3B & INT8 & 2.38\,GB & 32\% & 0.191 & 5.29$\times$ slower \\
NLLB-1.3B & INT4 & 1.78\,GB & 49\% & 0.054 & 1.51$\times$ slower \\
\bottomrule
\end{tabular}
\caption{Memory and latency across precisions, averaged over the four
Hausa/Swahili directions. Aya-23-8B at FP16 is sharded across two T4s;
peak memory is summed across devices.}
\label{tab:cost}
\end{table}

Two patterns stand out. First, memory savings are substantial and
reliable: 32--42\% at INT8 and 49--61\% at INT4, scaling with model
size. For Aya-23-8B specifically, INT4 reduces peak memory from
17.1\,GB to 6.6\,GB, moving the model from ``requires server-class
VRAM or multi-GPU sharding'' to ``fits comfortably on a single consumer
GPU.''

Second, latency does \emph{not} decrease---all quantised configurations
are slower than FP16. INT4 slowdowns are modest and comparable across
models (1.29$\times$--1.51$\times$), consistent with a per-matmul
dequantisation cost that dominates the small compute budget of each T4
operation. INT8 slowdowns are more severe and more variable:
1.93$\times$ for Aya-23-8B, 4.13$\times$ for Qwen-1.5B, and
5.29$\times$ for NLLB-1.3B. The smaller models are hit hardest in
relative terms, which we attribute to their FP16 baselines already
being latency-bound by the per-call overhead of the generation loop
rather than by compute, so that added dequantisation cost represents a
larger fractional increase.

This pattern---memory savings without latency savings on consumer
GPUs---undermines a common assumption that post-training quantisation
delivers both. For the deployment contexts motivating this work,
pre-packed-weight schemes \cite{frantar2023gptq,lin2024awq} or
kernel-level INT8 implementations are more promising than
bitsandbytes-style dynamic dequantisation, and are a natural extension
of this study.

\subsection{Resource gradient across languages}

For NLLB-1.3B at FP16, chrF on X$\rightarrow$English descends
monotonically: Swahili (63.77) $>$ Hausa (56.08) $>$ Yoruba (44.55).
The same ordering holds in the eng$\rightarrow$X direction (60.06,
51.84, 25.67). This gradient tracks reported training-data availability
for these languages in the NLLB-200 training corpus
\cite{nllb2024nature} and is a useful axis for subsequent compression
research: compression effects at the ``head'' of the gradient
(Swahili) may differ qualitatively from the ``tail'' (Yoruba), where
further analysis is needed.

\subsection{Model-specific observations}

\paragraph{Aya-23-8B over-generates.}
The hypothesis-to-reference length ratio for Aya at FP16 ranges from
1.18 (swh$\rightarrow$eng) to 1.88 (eng$\rightarrow$swh). Compared with
NLLB's near-unity ratios (0.96--0.99), Aya systematically produces
longer outputs than the reference, consistent with its
instruction-tuned behaviour not being adapted to concise translation
output. This partially explains Aya's lower chrF beyond raw translation
quality.

\paragraph{Qwen tokenisation of African orthographic markers.}
Qwen2.5's byte-level BPE encodes the Hausa hooked consonants---U+0253
(b with hook), U+0257 (d with hook), U+0199 (k with hook), and U+01B4
(y with hook)---as multi-byte sequences, which are often fragmented
into several subword pieces. Preliminary inspection suggests this may
contribute to Qwen's weak eng$\rightarrow$hau performance. We did not
run a controlled tokenisation study, and we flag this as a hypothesis
rather than a finding.

\section{Limitations and Threats to Validity}
\label{sec:limits}

\paragraph{(L1) Compression methods evaluated are limited.}
We test only bitsandbytes INT8 and NF4 INT4. GPTQ, AWQ, SmoothQuant
\cite{xiao2023smoothquant}, and distillation are absent, and the null
quality-degradation finding may not generalise.

\paragraph{(L2) Only one benchmark.}
FLORES-200 devtest is a single-domain, sentence-level, written-language
benchmark. Behaviour on noisier real-world text (social media, speech
transcripts) may differ.

\paragraph{(L3) No statistical significance testing was performed.}
Decoding is deterministic (greedy, beam size 1), so repeated runs of a
given configuration produce identical output; the relevant uncertainty
is sampling variance over the 1{,}012-sentence test set. Quantifying it
requires a paired bootstrap over sentences, which we did not run. We
therefore treat every delta in Table~\ref{tab:delta} as descriptive
only, and draw no conclusion from differences below roughly 1 chrF.

\paragraph{(L4) Latency measurements are on a specific hardware
configuration.}
Our T4 results reflect consumer-GPU behaviour but do not generalise to
A100/H100-class accelerators, where quantised kernels are better
optimised and may recover latency benefits. Latency is wall-clock time
per sentence at the batch sizes stated in the released results, and is
therefore a throughput-oriented rather than single-stream measurement.

\paragraph{(L5) Yoruba sweep incomplete.}
Only FP16 numbers are available for Yoruba, due to Kaggle session-time
limits during the Aya-23-8B runs. INT8 and INT4 Yoruba numbers are
future work, and the Yoruba rows were produced in a separate session
from the main sweep.

\paragraph{(L6) Per-sentence outputs are not released.}
We release aggregate metrics per configuration but not the underlying
hypothesis files, which limits independent re-scoring and error
analysis. Releasing them is a priority for the next revision.

\paragraph{(L7) Single researcher, no independent replication.}
The pilot was conducted by one researcher over eight weeks, and
individual runs were not independently replicated. Partial mitigations:
all code, hyperparameters, and library versions are released; FLORES is
a public benchmark; and metrics follow SacreBLEU standardised
signatures.

\section{Discussion and Research Directions}

Three findings motivate specific follow-on research.

\paragraph{On the LLM-versus-specialised-MT gap.}
The 33--43 chrF gap between NLLB-1.3B and the larger Aya-23-8B on
Hausa and Swahili suggests that at current scales and training recipes,
dedicated MT models remain decisively stronger for genuinely
low-resource language pairs. Research on ``efficient low-resource
translation'' should therefore prioritise making specialised MT models
more efficient and adaptable, rather than expecting general LLMs to
close the gap through scaling or compression alone. We note the
comparison is not scale-controlled: Aya-23 does not list these
languages among its 23 \cite{aryabumi2024aya23}, so the result
separates targeted training data from parameter count, not architecture
from architecture.

\paragraph{On the aggregate-quality finding.}
Null results in aggregate chrF should prompt---not foreclose---
finer-grained analysis. Specific linguistic phenomena (rare vocabulary,
morphology, named entities, numerals, non-ASCII orthography) may
degrade under compression even when overall chrF does not. Error-type
analysis across compression regimes is a direct and productive
follow-on, and requires the per-sentence outputs noted in L6.

\paragraph{On the latency finding.}
Compression methods designed with consumer-GPU deployment as a
first-class constraint---including pre-packed-weight quantisation,
hardware-aware kernel implementations, and selective-precision
schemes---are a natural research direction, particularly for the
low-resource contexts where consumer GPUs are most common.

\section*{Acknowledgements}

Compute: Kaggle (NVIDIA T4). Benchmark: FLORES-200. Models: Alibaba
Qwen team, Cohere For AI (Aya), Meta AI (NLLB).


\appendix

\section{Per-direction detailed results}

Table~\ref{tab:appendix} reports every metric for every evaluated
configuration. Blank cells (---) indicate metrics not recorded in the
Yoruba extension run, which logged a reduced set of fields. The
complete machine-readable results, including per-run library versions
and hardware metadata, are in \texttt{results/combined\_all.csv} in the
repository.

{\scriptsize
\input{appendix_a}
}

\end{document}

%% file: appendix_a.tex
\begin{longtable}{llcrrrrrr}
\caption{Full per-direction results for all evaluated configurations on FLORES-200 devtest (1{,}012 sentences per direction). ``Len.\ ratio'' is hypothesis-to-reference length ratio; ``s/sent'' is wall-clock inference time per sentence. Dashes mark metrics not recorded by the Yoruba extension run.}\label{tab:appendix} \\
\toprule
Model & Prec. & Direction & BLEU & spBLEU & chrF & chrF++ & Len. ratio & s/sent \\
\midrule
\endfirsthead
\multicolumn{9}{l}{\footnotesize\itshape Table \thetable\ continued from previous page.} \\
\toprule
Model & Prec. & Direction & BLEU & spBLEU & chrF & chrF++ & Len. ratio & s/sent \\
\midrule
\endhead
\midrule
\multicolumn{9}{r}{\footnotesize\itshape Continued on next page.} \\
\endfoot
\bottomrule
\endlastfoot
Qwen2.5-1.5B & FP16 & hau$\rightarrow$eng & 3.30 & 4.96 & 23.28 & 20.98 & 1.038 & 0.085 \\
 & FP16 & eng$\rightarrow$hau & 1.09 & 1.56 & 16.85 & 14.59 & 0.846 & 0.163 \\
 & FP16 & swh$\rightarrow$eng & 4.02 & 5.23 & 24.19 & 22.00 & 0.983 & 0.080 \\
 & FP16 & eng$\rightarrow$swh & 1.97 & 3.05 & 24.39 & 21.55 & 1.132 & 0.167 \\
 & FP16 & yor$\rightarrow$eng & 2.82 & -- & 22.90 & -- & -- & 0.402 \\
 & FP16 & eng$\rightarrow$yor & 0.67 & -- & 12.74 & -- & -- & 1.320 \\
 & INT8 & hau$\rightarrow$eng & 3.55 & 5.20 & 23.33 & 21.01 & 1.032 & 0.319 \\
 & INT8 & eng$\rightarrow$hau & 1.06 & 1.83 & 18.91 & 16.31 & 0.858 & 0.680 \\
 & INT8 & swh$\rightarrow$eng & 3.99 & 5.24 & 24.16 & 21.93 & 0.972 & 0.320 \\
 & INT8 & eng$\rightarrow$swh & 1.71 & 2.82 & 24.09 & 21.21 & 1.129 & 0.725 \\
 & INT4 & hau$\rightarrow$eng & 3.29 & 4.97 & 22.49 & 20.25 & 1.061 & 0.113 \\
 & INT4 & eng$\rightarrow$hau & 1.35 & 1.91 & 17.33 & 15.06 & 0.811 & 0.206 \\
 & INT4 & swh$\rightarrow$eng & 3.66 & 4.87 & 23.46 & 21.24 & 1.008 & 0.113 \\
 & INT4 & eng$\rightarrow$swh & 2.05 & 3.10 & 23.34 & 20.57 & 1.093 & 0.212 \\
\midrule
Aya-23-8B & FP16 & hau$\rightarrow$eng & 3.27 & 4.83 & 23.35 & 21.15 & 1.246 & 0.504 \\
 & FP16 & eng$\rightarrow$hau & 1.11 & 1.86 & 15.48 & 13.34 & 1.487 & 0.828 \\
 & FP16 & swh$\rightarrow$eng & 6.92 & 8.84 & 29.73 & 27.86 & 1.184 & 0.486 \\
 & FP16 & eng$\rightarrow$swh & 1.06 & 1.60 & 16.73 & 14.41 & 1.876 & 0.818 \\
 & FP16 & yor$\rightarrow$eng & 2.68 & -- & 19.88 & -- & -- & 1.499 \\
 & FP16 & eng$\rightarrow$yor & 0.20 & -- & 5.27 & -- & -- & 2.851 \\
 & INT8 & hau$\rightarrow$eng & 3.66 & 5.22 & 23.12 & 21.01 & 1.219 & 0.998 \\
 & INT8 & eng$\rightarrow$hau & 1.15 & 1.89 & 15.15 & 13.04 & 1.479 & 1.593 \\
 & INT8 & swh$\rightarrow$eng & 6.66 & 8.51 & 29.45 & 27.57 & 1.173 & 0.897 \\
 & INT8 & eng$\rightarrow$swh & 1.06 & 1.66 & 16.53 & 14.30 & 1.848 & 1.597 \\
 & INT4 & hau$\rightarrow$eng & 2.82 & 4.55 & 22.23 & 20.07 & 1.205 & 0.652 \\
 & INT4 & eng$\rightarrow$hau & 1.29 & 1.99 & 15.61 & 13.55 & 1.386 & 1.062 \\
 & INT4 & swh$\rightarrow$eng & 6.20 & 7.97 & 27.54 & 25.68 & 1.145 & 0.624 \\
 & INT4 & eng$\rightarrow$swh & 1.13 & 1.63 & 16.15 & 13.99 & 1.801 & 1.062 \\
\midrule
NLLB-1.3B & FP16 & hau$\rightarrow$eng & 33.22 & 35.08 & 56.08 & 54.69 & 0.963 & 0.032 \\
 & FP16 & eng$\rightarrow$hau & 24.23 & 26.46 & 51.84 & 49.38 & 0.959 & 0.042 \\
 & FP16 & swh$\rightarrow$eng & 41.63 & 43.37 & 63.77 & 62.46 & 0.963 & 0.032 \\
 & FP16 & eng$\rightarrow$swh & 31.41 & 35.19 & 60.06 & 57.53 & 0.991 & 0.038 \\
 & FP16 & yor$\rightarrow$eng & 20.54 & -- & 44.55 & -- & -- & 0.152 \\
 & FP16 & eng$\rightarrow$yor & 4.97 & -- & 25.67 & -- & -- & 0.303 \\
 & INT8 & hau$\rightarrow$eng & 33.26 & 35.11 & 56.11 & 54.70 & 0.964 & 0.170 \\
 & INT8 & eng$\rightarrow$hau & 24.26 & 26.52 & 51.94 & 49.47 & 0.961 & 0.222 \\
 & INT8 & swh$\rightarrow$eng & 41.65 & 43.38 & 63.78 & 62.47 & 0.962 & 0.168 \\
 & INT8 & eng$\rightarrow$swh & 31.29 & 35.06 & 59.93 & 57.40 & 0.992 & 0.202 \\
 & INT4 & hau$\rightarrow$eng & 33.60 & 35.21 & 56.08 & 54.73 & 0.971 & 0.050 \\
 & INT4 & eng$\rightarrow$hau & 24.09 & 26.23 & 51.80 & 49.32 & 0.957 & 0.061 \\
 & INT4 & swh$\rightarrow$eng & 41.69 & 43.23 & 63.66 & 62.36 & 0.963 & 0.049 \\
 & INT4 & eng$\rightarrow$swh & 30.82 & 34.25 & 59.29 & 56.76 & 0.985 & 0.057 \\
\end{longtable}